\documentclass[11pt]{article}
\usepackage[final]{acl}

\usepackage{times}
\usepackage{latexsym}
\usepackage[T1]{fontenc}
\usepackage[utf8]{inputenc}
\usepackage{microtype}
\usepackage{inconsolata}

\usepackage{graphicx}
\usepackage{booktabs}
\usepackage{amsmath}
\usepackage{amssymb}
\usepackage{xcolor}
\usepackage{multirow}
\usepackage{enumitem}
\usepackage{subcaption}
\usepackage{tikz}
\usetikzlibrary{shapes.geometric, arrows.meta, positioning, calc, fit, backgrounds}
\usepackage{pgfplots}
\pgfplotsset{compat=1.17}
\usepackage{listings}
\usepackage{tcolorbox}
\usepackage{adjustbox}
\usepackage{caption}
\usepackage{url}

\definecolor{rsatblue}{HTML}{4C72B0}
\definecolor{rsatred}{HTML}{C44E52}
\definecolor{rsatorange}{HTML}{DD8452}
\definecolor{rsatgray}{HTML}{999999}
\definecolor{codebg}{HTML}{F5F5F5}
\definecolor{citegreen}{HTML}{2E7D32}
\definecolor{citered}{HTML}{C62828}

\title{RSAT: Structured Attribution Makes Small Language Models Faithful Table Reasoners}

\author{
  Jugal Gajjar\thanks{Corresponding author.},\quad Kamalasankari Subramaniakuppusamy \\
  Department of Computer Science,\\The George Washington University, USA \\
  \texttt{\{jugal.gajjar, kamalasankaris\}@gwu.edu}
}

\begin{document}
\maketitle

\begin{abstract}
When a language model answers a table question, users have no way to verify which cells informed which reasoning steps. We introduce \textbf{RSAT}, a method that trains small language models (SLMs, 1--8B) to produce step-by-step reasoning with cell-level citations grounded in table evidence. Phase~1 (SFT) teaches a structured JSON output format from verified reasoning traces. Phase~2 (GRPO) optimizes a composite reward centered on NLI-based faithfulness, alongside citation validity and parsimony. Across six models from two families---Qwen~2.5 (1.5B/3B/7B) and Llama~3 (1B/3B/8B)---RSAT improves faithfulness \textbf{3.7$\times$} over SFT alone (0.224$\rightarrow$0.826), with near-perfect citation validity (0.992). Post-hoc attribution collapses below 13\% format success, confirming that attribution must be integrated into reasoning, not retrofitted. Ablations show the faithfulness reward is essential: removing it drops faithfulness from 0.97 to 0.03.
\end{abstract}

\section{Introduction}
\label{sec:intro}

When a language model answers a question over a table, users receive an answer with no way to verify \emph{which cells informed which reasoning steps}. This lack of interpretable, cell-level attribution limits trust in table reasoning systems, particularly in high-stakes domains such as financial analysis, journalism, and clinical decision support. While existing methods achieve strong accuracy---TAPAS \cite{herzig2020tapas} through cell selection, PASTA \cite{gu2022pasta} through operation-aware pre-training, chain-of-thought prompting \cite{wei2022chain} through step-by-step reasoning---none produce structured reasoning traces where every claim is explicitly grounded in specific table cells. The dominant alternative, \emph{post-hoc attribution}, where a model first generates an answer and then retroactively cites evidence, is appealing in principle but, as we demonstrate, collapses catastrophically on sub-8B models.

We propose \textbf{RSAT} (\textbf{R}einforcement-driven \textbf{S}tructured \textbf{A}ttribution \textbf{T}raining), a method that trains small language models (1--8B parameters) to produce structured JSON output where each reasoning step explicitly cites the table cells it depends on using $[\text{row}, \text{col}]$ coordinates. RSAT operates in two phases. \textbf{Phase~1 (SFT)} fine-tunes on 1{,}000 verified reasoning traces to teach the structured output format---achieving near-perfect format compliance but only ${\sim}$22\% faithfulness. \textbf{Phase~2 (GRPO)} applies group-relative reinforcement learning \cite{shao2024deepseekmath} with a composite reward that incentivizes answer correctness, citation validity, \emph{faithfulness}, and parsimony. The faithfulness signal---NLI entailment between cited cell values and reasoning text---is the key innovation: it teaches the model to \emph{ground} its reasoning in actual table evidence rather than producing structurally valid but unsubstantiated citations. GRPO makes this feasible on a single GPU by eliminating the critic model required by PPO, enabling composite reward optimization over non-differentiable signals like NLI scores.

We evaluate RSAT across six models from two architecture families---Qwen~2.5 (1.5B/3B/7B) and Llama~3 (1B/3B/8B)---on three table reasoning benchmarks (WTQ, FeTaQA, TabFact). Our experiments yield four key findings:

\begin{enumerate}[nosep,leftmargin=*]
\item RSAT improves faithfulness by 3.7$\times$ over SFT alone (0.224$\rightarrow$0.826 average across all models), while maintaining near-perfect citation validity (0.992) and format compliance (0.993).
\item Post-hoc attribution collapses entirely on some models (0.4\% format success for Qwen~3B), averaging under 13\% across all six---demonstrating that attribution must be integrated into reasoning, not retrofitted.
\item SFT and GRPO serve distinct, complementary roles: SFT teaches \emph{structure} ($+$0.61 format, $+$0.63 citation validity), while GRPO teaches \emph{quality} ($+$0.60 faithfulness). Neither alone is sufficient.
\item Ablations reveal faithfulness reward is the sole essential signal---removing it collapses faithfulness from 0.97 to 0.03, while removing other components causes only modest degradation.
\end{enumerate}

\noindent The complete code and data for RSAT are available at \url{https://github.com/JugalGajjar/RSAT}.

\section{Related Work}
\label{sec:related}
 
\paragraph{Table Question Answering.}
Table QA has progressed from semantic parsing \cite{pasupat2015compositional,zhong2017seq2sql} to specialized pre-trained architectures: TAPAS \cite{herzig2020tapas} extends BERT \cite{devlin2019bert} with row and column positional embeddings, TAPEX \cite{liu2021tapex} pre-trains by mimicking an SQL executor, and TaBERT \cite{yin2020tabert} jointly pre-trains on text-table pairs. More recent work leverages LLM prompting---Binder \cite{cheng2022binding} maps questions to executable SQL or Python, Chain-of-Table \cite{wang2024chain} iteratively transforms tables within the reasoning chain, and TaPERA \cite{zhao2024tapera} decomposes questions into sub-questions answered by executable programs. Despite strong accuracy gains, none of these methods produce step-by-step reasoning with explicit cell-level attribution.
 
\paragraph{Table Fact Verification.}
TabFact \cite{chen2019tabfact} established table-based fact verification as a binary classification task. PASTA \cite{gu2022pasta} designs operation-aware cloze pre-training to teach table operations, achieving strong accuracy with DeBERTaV3. Other approaches use program-guided reasoning \cite{yang2020program} or joint verification and retrieval \cite{schlichtkrull2021joint,eisenschlos2020understanding}. All output binary entailed/refuted labels without explaining \emph{which cells} support the verdict.
 
\paragraph{Attribution and Faithfulness in LLMs.}
\citet{rashkin2023measuring} formalize attributable generation, defining metrics for whether outputs are supported by cited sources. Self-RAG \cite{asai2023self} trains models to generate self-reflection tokens assessing retrieval relevance and faithfulness during generation. ALCE \cite{gao2023enabling} benchmarks citation generation for long-form text QA. RARR \cite{gao2023rarr} retrofits attributions post-hoc. \citet{bohnet2022attributed} study attributed QA with retrieval-augmented approaches. \citet{wallat2024correctness} argue that correctness alone is insufficient---citations must also be faithful to the model's actual reasoning, not post-rationalized. All these works operate on unstructured text; RSAT extends attribution to \emph{structured table evidence} with cell-level granularity.
 
\paragraph{RL for LLM Reasoning.}
RLHF \cite{ouyang2022training} aligns models with human preferences via learned reward models, while DPO \cite{rafailov2023direct} removes the reward model entirely. DeepSeek-R1 \cite{guo2025deepseek} demonstrates that GRPO \cite{shao2024deepseekmath}---which normalizes rewards within candidate groups, eliminating the critic model---can incentivize chain-of-thought reasoning. \citet{dang2025reinforcement} show GRPO enhances mathematical reasoning in 1.5B models with modest compute. We design a domain-specific composite reward for table attribution quality---combining NLI-based faithfulness, citation validity, and parsimony---and optimize it with GRPO. To our knowledge, RSAT is the first application of RL-driven attribution training to structured table reasoning.
 
\paragraph{Positioning.}
Existing table QA methods optimize accuracy without interpretable attribution. Fact verification outputs binary labels without reasoning traces. Text-domain attribution methods like Self-RAG operate on passage-level evidence, not cell-level structure. RSAT uniquely combines structured reasoning, cell-level citations, NLI-based faithfulness, and RL-driven attribution training in small (1--8B) models.

\section{Methodology}
\label{sec:method}

\begin{figure*}[t]
\centering
\includegraphics[width=\textwidth]{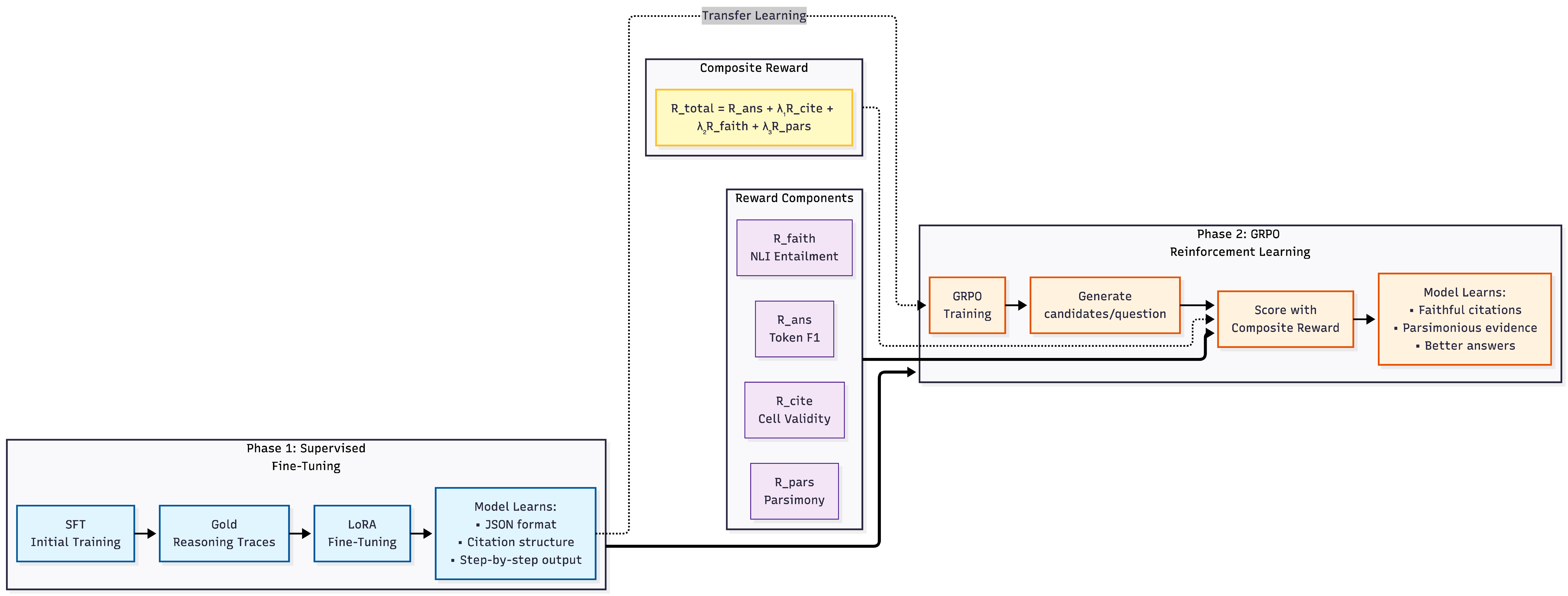}
\caption{\textbf{RSAT overview.} Phase~1 (SFT) teaches the structured output format using verified gold traces. Phase~2 (GRPO) generates multiple candidates per question, scores them with a composite reward, and updates the policy. The faithfulness reward is the critical signal that grounds reasoning in actual table evidence.}
\label{fig:overview}
\end{figure*}

\subsection{Task Formulation}
\label{sec:task}

Given a table $T$ with headers $\{h_1, \ldots, h_C\}$ and rows $\{r_1, \ldots, r_R\}$, and a natural language question~$q$, the model must produce a structured JSON output containing: (1) a list of \texttt{reasoning\_steps}, where each step is a natural language claim paired with \texttt{cited\_cells}---a list of $[\text{row}_i, \text{col}_j]$ coordinates referencing specific table cells that support the claim; and (2) a final \texttt{answer} string. Cell coordinates are zero-indexed: $[0, 0]$ denotes the first data cell (row~0, column~0), excluding the header row. For example, a reasoning step might be: \texttt{\{"step": "Team A had 15 wins", "cited\_cells": [[2,3], [0,3]]\}}, where cells $[2,3]$ and $[0,3]$ contain the relevant values from the wins column. This format enables post-hoc \emph{auditing}: a user can independently verify each claim by checking the cited cells against the original table.

Tables are serialized into a flat text format that preserves structural information: \texttt{[HEADER] col1 | col2 | ... [ROW~0] val1 | val2 | ... [ROW~1] ...}. The row index markers allow the model to learn a direct mapping between cell references in its output and positions in the serialized input.

\subsection{Data Construction}
\label{sec:data}

We draw from three table reasoning benchmarks: WikiTableQuestions (WTQ) \cite{pasupat2015compositional} for factoid QA, FeTaQA \cite{nan2022fetaqa} for free-form long-answer QA, and TabFact \cite{chen2019tabfact} for fact verification. Table~\ref{tab:data_stats} summarizes the full corpus.

\begin{table}[t]
\centering
\caption{\textbf{Dataset statistics.} SFT examples include teacher-generated reasoning traces; GRPO examples contain only table-question-answer triples.}
\label{tab:data_stats}
\small
\setlength{\tabcolsep}{3.5pt}
\begin{tabular}{@{}l rrr r@{}}
\toprule
\textbf{Split} & \textbf{WTQ} & \textbf{FeTaQA} & \textbf{TabFact} & \textbf{Total} \\
\midrule
SFT train      & 630    & 135    & 135    & 900 \\
SFT val        & 70     & 15     & 15     & 100 \\
\midrule
GRPO train     &        &        &        & 38{,}647 \\
GRPO val       & \multicolumn{3}{c}{(combined pool)} & 16{,}624 \\
GRPO test      &        &        &        & 19{,}126 \\
\midrule
\textbf{Total corpus} & & & & \textbf{75{,}397} \\
\bottomrule
\end{tabular}
\end{table}

\paragraph{SFT data (1{,}000 examples).}
For each example, we prompt Claude Opus 4.5 to generate a structured reasoning trace given the serialized table and question. Each trace is programmatically verified for (i) JSON validity, (ii) cell coordinate bounds, and (iii) step count (3--4 required). Failed traces are re-generated in batched repair passes with error-specific prompts. Notably, the verification pipeline enforces structural correctness (valid JSON, valid cell coordinates, and step count) but does not guarantee semantic grounding between cited cells and reasoning steps. This explains why SFT achieves near-perfect format compliance but relatively low faithfulness ($\sim$22\%): the model learns to imitate structure without necessarily grounding its reasoning in the cited evidence. The average verified trace contains 3.2 steps with 2.4 cited cells per step. The set is split 900/100 (train/val), stratified by source.

\paragraph{GRPO data (74{,}397 examples).}
A larger pool of table-question-answer triples (no reasoning traces) is drawn from the same benchmarks. The model generates its own candidates during RL; only gold answers are needed for reward computation. We subsample 500 training and 500 test examples per model to keep compute tractable.

\subsection{Phase 1: Supervised Fine-Tuning}
\label{sec:sft}

The SFT phase teaches the model the structured output format. We apply LoRA \cite{hu2022lora} adapters to all linear projection layers (Q, K, V, O, gate, up, down) and fine-tune for 3 epochs with a cosine learning rate schedule.

SFT achieves ${\sim}99\%$ format success and ${\sim}99\%$ citation validity across all models---but only ${\sim}22\%$ faithfulness on average. This gap is the central motivation for Phase~2: the model learns \emph{what} to produce (valid JSON with cell references) but not \emph{how well} to ground its reasoning in the cited evidence.

\subsection{Phase 2: Group Relative Policy Optimization}
\label{sec:grpo}

To improve attribution quality beyond supervised learning, we apply GRPO \cite{shao2024deepseekmath,guo2025deepseek}. We prefer GRPO over PPO \cite{schulman2017proximal} because it eliminates the critic model, reducing memory overhead by roughly half---critical for single-GPU training. We prefer it over DPO \cite{rafailov2023direct} because our multi-component reward requires scoring individual outputs, not ranking pairs.

For each training question, the model generates $G{=}8$ candidate, where each candidate is scored by the composite reward function (Section~\ref{sec:reward}). Advantages are computed as group-relative z-scores---each candidate's reward minus the group mean, divided by the group standard deviation. The policy is updated via the clipped surrogate objective. We attach a fresh LoRA adapter to the merged SFT checkpoint and train for one epoch over 500 subsampled training examples (250 optimization steps).

\subsection{Composite Reward Function}
\label{sec:reward}

The composite reward balances four objectives plus a hard format gate:
\begin{equation}
R = R_{\text{ans}} + \lambda_1 R_{\text{cite}} + \lambda_2 R_{\text{faith}} + \lambda_3 R_{\text{pars}} + R_{\text{fmt}}
\label{eq:reward}
\end{equation}

\noindent\textbf{Answer reward} ($R_{\text{ans}}$): Token-level F1 between the predicted and gold answer. When multiple gold answers exist, we take the maximum.

\noindent\textbf{Citation validity} ($R_{\text{cite}}$, $\lambda_1{=}0.3$): Fraction of cited cell coordinates $[r_i, c_j]$ that fall within the table's dimensions ($0 \leq r_i < R$, $0 \leq c_j < C$). Measures structural correctness of citations.

\noindent\textbf{Faithfulness} ($R_{\text{faith}}$, $\lambda_2{=}0.5$): For each reasoning step, we concatenate the values of the cited cells into an evidence string and compute the entailment probability against the step's text using DeBERTa-v3-base NLI \cite{he2020deberta}. The score is averaged across all steps; steps citing no cells receive a faithfulness score of 0. This is the core attribution signal---it measures whether the cited cells actually \emph{support} the claim, not merely whether the citations are structurally valid.

\noindent\textbf{Parsimony} ($R_{\text{pars}}$, $\lambda_3{=}0.2$): Penalizes over-citation. Steps citing $\leq 3$ cells receive full score (1.0); the score decays linearly to 0.0 at $\geq 8$ cells. This prevents reward hacking through exhaustive citation of all cells.

\noindent\textbf{Format penalty} ($R_{\text{fmt}}$): Acts as a hard gate: $-1$ if the output fails JSON parsing, $0$ otherwise. When triggered, this penalty dominates all other components, ensuring the model maintains the structured format learned during SFT.

The weighting hierarchy reflects our design priorities: faithfulness is the primary objective, citation validity is a structural prerequisite, and parsimony is a refinement. We set weights based on preliminary sensitivity analysis on Qwen 7B, tracking faithfulness, answer F1, and training stability across configurations. We found that faithfulness was most sensitive to $\lambda_2$, while $\lambda_1$ and $\lambda_3$ primarily influenced structural correctness and verbosity. The final configuration ($\lambda_2 > \lambda_1 > \lambda_3$) reflects this hierarchy, prioritizing grounding quality while maintaining valid and concise citations. Results were stable under $\pm$0.1 perturbations. Full mathematical definitions are in Appendix~\ref{app:reward}.

\section{Experimental Setup}
\label{sec:setup}

\paragraph{Models.}
We evaluate two model families across three scales: \textbf{Qwen~2.5 Instruct} (1.5B, 3B, 7B) \cite{qwen2024qwen25} and \textbf{Llama~3 Instruct} (1B, 3B, 8B) \cite{llama2024llama3}. All models are trained with LoRA \cite{hu2022lora} in bf16 precision on a single NVIDIA H100 80GB GPU. Total compute is approximately 18.4 GPU-hours for the six main models, plus 18.3 GPU-hours for ablation runs---36.8 GPU-hours in total (Appendix~\ref{app:compute}).

\paragraph{Baselines.}
We compare four methods applied to each model:
(1)~\textbf{Zero-shot}: the base instruction-tuned model prompted with the RSAT system prompt, no fine-tuning;
(2)~\textbf{SFT-only}: Phase~1 only (no GRPO);
(3)~\textbf{Post-hoc}: a two-pass baseline where the model first generates a chain-of-thought answer, then is prompted in a second pass to map its reasoning to cell coordinates in JSON format (using the same system prompt as RSAT to ensure a fair comparison);
(4)~\textbf{RSAT}: the full two-phase pipeline.

\paragraph{Evaluation.}
We evaluate on 500 held-out examples from the combined WTQ/FeTaQA/TabFact test split using greedy decoding. We report six metrics:
\textbf{Faithfulness} (NLI entailment between cited cells and reasoning text; \emph{primary metric}),
\textbf{Answer F1} (token-level overlap with gold answer), 
\textbf{Citation Validity} (fraction of cited cells within table bounds),
\textbf{Parsimony} (penalizes over-citation; see Section~\ref{sec:reward}), 
\textbf{Format Success} (fraction of outputs parsing as valid JSON), and
\textbf{Answer EM} (exact string match; reported for completeness but expected to be low due to answer paraphrasing in table QA).

\paragraph{Training Details.}
\textbf{SFT:} LoRA rank 32--64, learning rate $2{\times}10^{-4}$, 3 epochs, effective batch size 16. 
\textbf{GRPO:} LoRA rank 16--32, learning rate $5{\times}10^{-5}$, 1 epoch over 500 examples, $G{=}8$, temperature~0.9. Faithfulness scoring uses DeBERTa-v3-base NLI (\texttt{cross-encoder/nli-deberta-v3-base}) during both training and evaluation, ensuring metric consistency. Implementation uses TRL~v0.29.0. 
Full hyperparameters are in Appendix~\ref{app:hyperparams}.

\paragraph{Robustness of Faithfulness Metric.}
Faithfulness is both the training reward and the primary evaluation metric, computed by the same DeBERTa-v3 NLI model. This introduces a potential train--evaluation circularity. Two observations partially address this concern. First, ablation results show that removing the faithfulness reward causes catastrophic collapse (0.97$\to$0.03) while leaving other metrics largely intact---a model purely gaming the proxy would not require the faithfulness signal so critically. Second, faithfulness improvements are accompanied by consistent answer F1 gains (+0.09 average), indicating that grounding improvements transfer to task performance. A human evaluation study remains the necessary next step to fully validate the NLI proxy.

\section{Results and Analysis}
\label{sec:results}

\subsection{Main Results}

Table~\ref{tab:main} presents results across all six models and four methods. RSAT achieves the best performance on every metric for every model. The most striking result is the faithfulness improvement: averaged across all six models, RSAT achieves 0.826 faithfulness compared to 0.224 for SFT-only---a \textbf{3.7$\times$} improvement. This confirms that the GRPO composite reward successfully teaches models to ground reasoning in actual table evidence.

Answer F1 also improves consistently across all models ($+$0.09 average over SFT), confirming that faithfulness training does not sacrifice answer quality. Citation validity and format success remain near-perfect ($>$0.97), indicating that GRPO preserves the structural gains from SFT.

\begin{table*}[t]
\centering
\caption{\textbf{Main results across six models and four methods.} Best values per model are \textbf{bolded}. RSAT achieves the highest scores on every metric for every model. Faithfulness shows the largest improvement: 3.7$\times$ over SFT on average. EM is omitted from the main table due to uniformly low values (0.000--0.018) caused by strict string matching; F1 captures partial correctness.}
\label{tab:main}
\small
\setlength{\tabcolsep}{4.5pt}
\begin{tabular}{@{}ll ccccc@{}}
\toprule
\textbf{Model} & \textbf{Method} & \textbf{F1}$\uparrow$ & \textbf{Cite}$\uparrow$ & \textbf{Faith}$\uparrow$ & \textbf{Pars}$\uparrow$ & \textbf{Fmt\%}$\uparrow$ \\
\midrule
\multirow{4}{*}{Qwen 1.5B}
 & Zero-Shot & 0.209 & 0.391 & 0.040 & 0.389 & 0.436 \\
 & SFT       & 0.371 & 0.995 & 0.149 & 0.918 & 0.998 \\
 & Post-Hoc  & 0.115 & 0.182 & 0.032 & 0.209 & 0.244 \\
 & \textbf{RSAT} & \textbf{0.524} & \textbf{0.996} & \textbf{0.847} & \textbf{0.990} & \textbf{0.998} \\
\midrule
\multirow{4}{*}{Qwen 3B}
 & Zero-Shot & 0.181 & 0.296 & 0.040 & 0.364 & 0.398 \\
 & SFT       & 0.531 & 0.996 & 0.213 & 0.848 & 0.998 \\
 & Post-Hoc  & 0.002 & 0.000 & 0.000 & 0.004 & 0.004 \\
 & \textbf{RSAT} & \textbf{0.592} & \textbf{0.999} & \textbf{0.946} & \textbf{0.996} & \textbf{1.000} \\
\midrule
\multirow{4}{*}{Qwen 7B}
 & Zero-Shot & 0.277 & 0.565 & 0.086 & 0.501 & 0.566 \\
 & SFT       & 0.576 & 1.000 & 0.234 & 0.888 & 1.000 \\
 & Post-Hoc  & 0.088 & 0.140 & 0.027 & 0.117 & 0.140 \\
 & \textbf{RSAT} & \textbf{0.619} & \textbf{0.992} & \textbf{0.977} & \textbf{0.992} & \textbf{0.992} \\
\midrule
\multirow{4}{*}{Llama 1B}
 & Zero-Shot & 0.100 & 0.326 & 0.008 & 0.280 & 0.334 \\
 & SFT       & 0.403 & 0.947 & 0.192 & 0.787 & 0.952 \\
 & Post-Hoc  & 0.066 & 0.064 & 0.004 & 0.184 & 0.198 \\
 & \textbf{RSAT} & \textbf{0.537} & \textbf{0.972} & \textbf{0.480} & \textbf{0.967} & \textbf{0.972} \\
\midrule
\multirow{4}{*}{Llama 3B}
 & Zero-Shot & 0.099 & 0.368 & 0.015 & 0.336 & 0.368 \\
 & SFT       & 0.546 & 0.984 & 0.269 & 0.822 & 0.984 \\
 & Post-Hoc  & 0.035 & 0.056 & 0.005 & 0.085 & 0.090 \\
 & \textbf{RSAT} & \textbf{0.592} & \textbf{0.993} & \textbf{0.735} & \textbf{0.865} & \textbf{0.994} \\
\midrule
\multirow{4}{*}{Llama 8B}
 & Zero-Shot & 0.054 & 0.164 & 0.006 & 0.142 & 0.164 \\
 & SFT       & 0.555 & 0.996 & 0.288 & 0.830 & 0.998 \\
 & Post-Hoc  & 0.013 & 0.040 & 0.006 & 0.038 & 0.040 \\
 & \textbf{RSAT} & \textbf{0.647} & \textbf{1.000} & \textbf{0.972} & \textbf{1.000} & \textbf{1.000} \\
\bottomrule
\end{tabular}
\end{table*}

The more modest gains in answer F1 relative to faithfulness reflect a difference in objectives rather than a limitation. RSAT targets verifiability---whether reasoning is auditably grounded in table evidence---rather than raw accuracy. In high-stakes domains such as financial analysis or clinical decision support, this distinction matters: a correct answer that cannot be traced to specific evidence offers users no basis for trust or audit.

\paragraph{Phase contribution.}
Table~\ref{tab:phase} quantifies what each training phase contributes on average across all six models. SFT produces the structural leap: format success jumps $+0.61$, citation validity $+0.64$. However, SFT yields only $+0.19$ average faithfulness---the model learns to produce well-formed JSON with valid cell references, but cited cells do not reliably support the reasoning claims. GRPO closes this gap, contributing $+0.60$ faithfulness while preserving structural quality established by SFT.

\begin{table}[t]
\centering
\caption{\textbf{Phase contribution analysis.} Average per-metric gain from each training stage across all six models. SFT drives structure; GRPO drives faithfulness.}
\label{tab:phase}
\small
\setlength{\tabcolsep}{5pt}
\begin{tabular}{@{}lcccc@{}}
\toprule
\textbf{Transition} & \textbf{$\Delta$Fmt} & \textbf{$\Delta$Cite} & \textbf{$\Delta$Faith} & \textbf{$\Delta$F1} \\
\midrule
Zero-shot $\to$ SFT  & $+0.61$ & $+0.64$ & $+0.19$ & $+0.34$ \\
SFT $\to$ RSAT       & $+0.00$ & $+0.01$ & $\mathbf{+0.60}$ & $+0.09$ \\
\bottomrule
\end{tabular}
\end{table}

\subsection{Scaling Analysis}

\begin{figure}[t]
\centering
\includegraphics[width=\columnwidth]{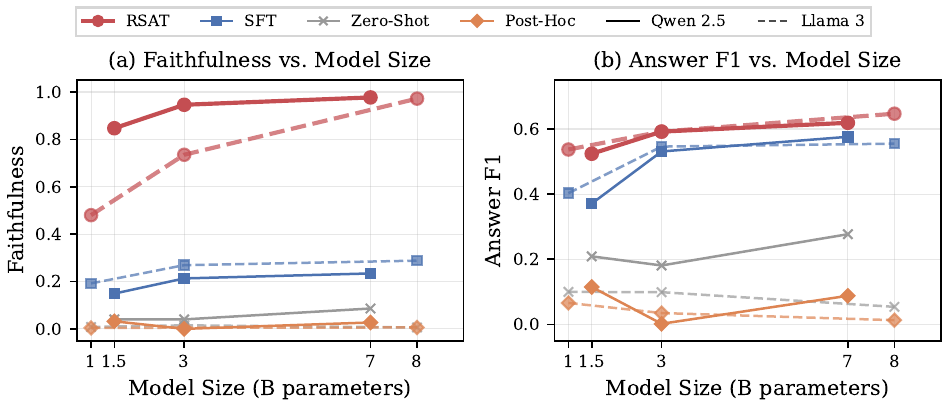}
\caption{\textbf{Scaling curves for faithfulness and answer F1.} RSAT (solid, high) dominates at every scale. Qwen outperforms Llama at small scale; the gap narrows at 7--8B. Post-hoc collapses regardless of scale.}
\label{fig:scaling}
\end{figure}

Figure~\ref{fig:scaling} presents faithfulness and answer F1 across model sizes for both families. Two patterns emerge.

First, \textbf{Qwen consistently outperforms Llama at equivalent scale on faithfulness}: 0.847 vs.\ 0.480 at ${\sim}1.5$B, 0.946 vs.\ 0.735 at 3B, converging at 7--8B (0.977 vs.\ 0.972). This gap may reflect differences in pre-training data composition or Qwen's larger vocabulary (151K vs.\ 128K tokens), which provides more expressive capacity for structured JSON output.

Second, \textbf{faithfulness scales with model size but with diminishing returns}. Qwen approaches ceiling ($>0.94$) at 3B, while Llama requires 8B to reach equivalent performance. This suggests a practical finding: for resource-constrained deployment, Qwen~3B offers the best faithfulness-per-parameter ratio.

One anomaly: Llama~3B shows lower parsimony (0.865) than both Llama~1B (0.967) and 8B (1.000), suggesting that mid-scale models may generate redundant reasoning steps that smaller and larger models avoid.

\subsection{Post-hoc Attribution Collapse}

\begin{figure}[t]
\centering
\includegraphics[width=\columnwidth]{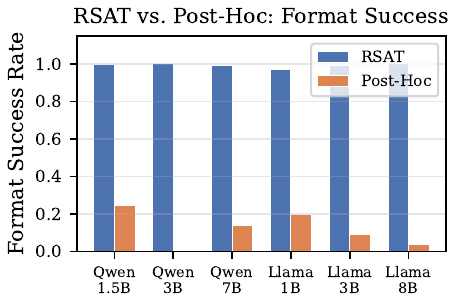}
\caption{\textbf{Post-hoc attribution collapses across all models.} RSAT maintains $>$97\% format success; post-hoc averages 12.7\% (dashed line).}
\label{fig:posthoc}
\end{figure}

The post-hoc baseline performs worse than every other method across all models---frequently worse than zero-shot. Average format success is just 12.7\%, compared to 99.3\% for RSAT (Figure~\ref{fig:posthoc}). The failure is particularly dramatic for Qwen~3B (0.4\%) and Llama~8B (4.0\%). The two-pass approach asks the model to retroactively map free-form reasoning to specific cell coordinates---a task requiring re-reading the table, identifying which cells correspond to which claims, and formatting the result as valid JSON. Small models lack the working memory for this mapping. In contrast, RSAT models produce citations \emph{during} reasoning, making attribution a natural part of the generation process.

Importantly, the dominant failure mode is not incorrect attribution but the failure to produce structured output at all. As shown in Appendix~\ref{app:posthoc_analysis}, 76--100\% of post-hoc outputs are empty or non-JSON, indicating that models consistently fail to comply with the required format in the second pass. This suggests a fundamental capacity limitation rather than a weak baseline design: small models lack the working memory to re-read the table and retroactively map free-form reasoning into structured citations. All methods use the same system prompt and base model, ensuring that the only difference is whether attribution is integrated during generation or applied post-hoc.

\subsection{Ablation Study}

Table~\ref{tab:ablation} shows the effect of removing individual reward components on Qwen~7B and Llama~8B.

\begin{table}[t]
\centering
\caption{\textbf{Ablation study.} Removing faithfulness reward causes catastrophic degradation ($-$86\% to $-$97\%). Removing parsimony causes significant over-citation. Citation reward has the smallest individual effect, as SFT already handles structural validity.}
\label{tab:ablation}
\small
\setlength{\tabcolsep}{4pt}
\begin{tabular}{@{}ll ccc@{}}
\toprule
\textbf{Model} & \textbf{Variant} & \textbf{F1} & \textbf{Faith} & \textbf{Pars} \\
\midrule
\multirow{4}{*}{\shortstack[l]{Qwen\\7B}}
 & RSAT (full)       & 0.619 & 0.977 & 0.992 \\
 & $-$Faithfulness   & 0.635 & \color{citered}{0.117} & 1.000 \\
 & $-$Parsimony      & 0.612 & 0.952 & \color{citered}{0.604} \\
 & $-$Citation       & 0.605 & 0.934 & 0.993 \\
\midrule
\multirow{4}{*}{\shortstack[l]{Llama\\8B}}
 & RSAT (full)       & 0.647 & 0.972 & 1.000 \\
 & $-$Faithfulness   & 0.638 & \color{citered}{0.031} & 0.996 \\
 & $-$Parsimony      & 0.626 & 0.899 & \color{citered}{0.575} \\
 & $-$Citation       & 0.617 & 0.938 & 1.000 \\
\bottomrule
\end{tabular}
\end{table}

\noindent\textbf{Faithfulness reward is critical.} Removing it causes faithfulness to plummet from 0.977 to 0.117 (Qwen~7B) and 0.972 to 0.031 (Llama~8B). Notably, answer F1 remains comparable or slightly improves, while parsimony and citation stay high. Without the faithfulness signal, the model learns to produce perfectly formatted, parsimonious outputs that cite real cells---but the citations are essentially \emph{random}. The model games the remaining rewards without grounding its reasoning. Interestingly, F1 slightly \emph{increases} without the faithfulness signal (0.619$\rightarrow$0.635 for Qwen~7B), revealing a mild tension: the model produces marginally better answers when freed from the constraint of justifying them with evidence.

\noindent\textbf{Parsimony reward prevents over-citation.} Without it, parsimony drops by ${\sim}40\%$ (0.992$\rightarrow$0.604 for Qwen, 1.000$\rightarrow$0.575 for Llama), and models default to citing 5--6 cells per step rather than 1--2. Faithfulness also degrades as over-citing dilutes evidence quality.

\noindent\textbf{Citation reward provides refinement.} Removing it causes only a ${\sim}3$--4\% faithfulness drop, because SFT already teaches valid cell coordinate production. The citation reward serves as a safety net, not a primary learning signal. See Appendix~\ref{app:ablation_viz} for visualizations.

\subsection{Qualitative Analysis}

\begin{figure}[t]
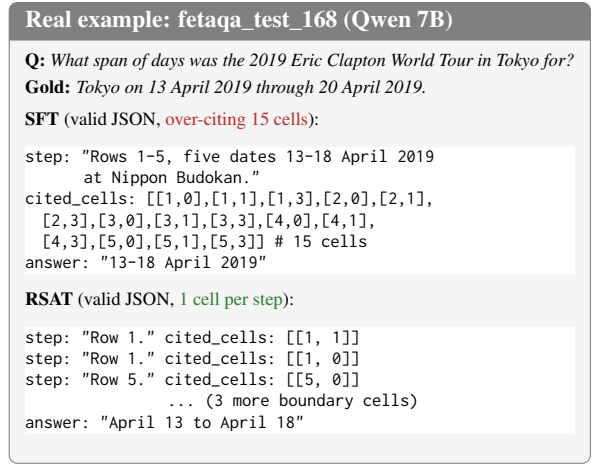

\begin{tcolorbox}[colback=codebg, colframe=gray!100, fonttitle=\bfseries\small,
  title={Real example: fetaqa\_test\_168 (Qwen~7B)}, boxrule=0.4pt,
  left=3pt, right=3pt, top=2pt, bottom=2pt]
\scriptsize
\textbf{Q:} \textit{What span of days was the 2019 Eric Clapton World Tour in Tokyo for?}\\[2pt]
\textbf{Gold:} \textit{Tokyo on 13 April 2019 through 20 April 2019.}\\[4pt]
\textbf{SFT} (valid JSON, \textcolor{citered}{over-citing 15 cells}):
\begin{lstlisting}[frame=none,backgroundcolor=\color{white},basicstyle=\ttfamily\scriptsize]
step: "Rows 1-5, five dates 13-18 April 2019
       at Nippon Budokan."
cited_cells: [[1,0],[1,1],[1,3],[2,0],[2,1],
  [2,3],[3,0],[3,1],[3,3],[4,0],[4,1],
  [4,3],[5,0],[5,1],[5,3]]   # 15 cells
answer: "13-18 April 2019"
\end{lstlisting}
\textbf{RSAT} (valid JSON, \textcolor{citegreen}{1 cell per step}):
\begin{lstlisting}[frame=none,backgroundcolor=\color{white},basicstyle=\ttfamily\scriptsize]
step: "Row 1."   cited_cells: [[1, 1]]
step: "Row 1."   cited_cells: [[1, 0]]
step: "Row 5."   cited_cells: [[5, 0]]
                 ... (3 more boundary cells)
answer: "April 13 to April 18"
\end{lstlisting}
\end{tcolorbox}
\caption{\textbf{Qualitative comparison (real test example).} SFT cites 15 cells in one step, bundling five complete table rows. RSAT cites one cell per step, pointing to exactly the evidence needed. (Both models produce slightly imprecise date ranges compared to the gold answer; the comparison illustrates attribution quality, not answer accuracy.)}
\label{fig:qualitative}
\end{figure}

Figure~\ref{fig:qualitative} shows a real test example comparing Qwen~7B SFT-only vs.\ RSAT on a FeTaQA concert tour question. The contrast reveals GRPO's learned citation behavior: SFT bundles the entire table section (15 cells across 5 rows) into a single step, while RSAT points to one cell per step---exactly the date cells needed to establish the span. Both outputs are valid JSON with correct answers; the difference is attribution quality and conciseness. RSAT completion length averages 71--211 tokens across models, compared to ${\sim}250$ tokens for SFT (Appendix~\ref{app:completion_length}).

\section{Discussion and Limitations}
\label{sec:discussion}

The central insight from our experiments is that format compliance and faithfulness are orthogonal capabilities in small language models. Supervised learning solves format almost completely (${\sim}$99\%) but barely touches faithfulness (${\sim}$22\%); only reward-driven optimization bridges the gap. Practically, \emph{high format compliance should not be mistaken for high attribution quality}.

\paragraph{Why SFT fails at faithfulness.}
We conjecture that SFT teaches surface-level imitation: the model copies the \emph{format} of gold traces---valid JSON, plausible cell references---without internalizing the semantic relationship between cited cells and claims. The 22\% faithfulness likely reflects the base rate at which randomly selected valid cells happen to entail the step text. GRPO's faithfulness reward provides the missing signal.

\paragraph{Practical lessons for GRPO training.}
For practitioners: \emph{monitor reward curves, not loss}. GRPO loss remained near zero throughout training for all six models, yet reward means consistently increased (${\sim}$0.75$\rightarrow$${\sim}$1.5) and eval metrics confirmed genuine learning. The near-zero loss is expected---GRPO normalizes advantages to zero mean within each group. Training curves are in Appendix~\ref{app:training_curves}.

\paragraph{Learned conciseness.}
The parsimony reward drives dramatic output compression: average completion length drops from ${\sim}$250 tokens (SFT) to 71--211 tokens (RSAT), with some models converging to 1.0 reasoning step per output. Whether this represents optimal conciseness or over-compression warrants further investigation.

\paragraph{Limitations.}
Several limitations should be noted.
(1)~\textbf{Train-eval circularity.} Faithfulness is both the training reward and the primary evaluation metric, both scored by the same DeBERTa NLI model. The model may learn to produce text that scores well on DeBERTa specifically rather than achieving genuine grounding. Additionally, if the scorer is noisy or biased toward certain linguistic patterns, the model may inherit these imperfections. A human evaluation study is needed to validate the NLI proxy.
(2)~\textbf{In-distribution evaluation.} All test examples come from WTQ, FeTaQA, and TabFact; generalization to unseen domains (e.g., financial, medical, or scientific tables) remains untested. These domains often feature more complex schemas and domain-specific semantics that may challenge the model's ability to ground reasoning accurately.
(3)~\textbf{Low exact match.} EM ranges from 0.000--0.018 across all experiments, reflecting strict string matching on paraphrased answers. F1 captures the partial correctness that EM misses, but this limits direct comparison with systems reporting only EM.

\paragraph{Future work.}
The most pressing next step is a human evaluation (50--100 examples) to validate NLI-based faithfulness and quantify the train-eval circularity gap. Replacing the NLI scorer with a learned reward model trained on human faithfulness judgments would break this circularity entirely. Cross-domain evaluation on financial or scientific tables would test generalization. More broadly, RSAT demonstrates that RL with composite, domain-specific rewards can teach small models capabilities that supervised learning alone cannot---a paradigm that may extend beyond tables to attributed reasoning over knowledge graphs, code, or APIs.

\section{Conclusion}
\label{sec:conclusion}

Faithful attribution in table reasoning is not a post-processing problem — it is a training objective. RSAT demonstrates this through a two-phase approach: SFT teaches structured output format, and GRPO with a composite reward teaches the model to \emph{ground} that structure in actual table evidence. Across six models from two architecture families, RSAT achieves 3.7$\times$ faithfulness improvement over SFT alone, with near-perfect citation validity — while post-hoc attribution collapses below 13\% format success. Ablations confirm that the NLI-based faithfulness reward is the sole essential signal.

Our results suggest a broader principle: for structured generation tasks where output \emph{format} is easy to learn but output \emph{quality} is not, supervised learning and reinforcement learning play complementary, non-substitutable roles. The complete code and data for RSAT are available at \url{https://github.com/JugalGajjar/RSAT}.

\bibliography{references}

@inproceedings{herzig2020tapas,
  title={TaPas: Weakly supervised table parsing via pre-training},
  author={Herzig, Jonathan and Nowak, Pawel Krzysztof and M{\"u}ller, Thomas and Piccinno, Francesco and Eisenschlos, Julian},
  booktitle={Proceedings of the 58th annual meeting of the association for computational linguistics},
  pages={4320--4333},
  year={2020}
}

@article{liu2021tapex,
  title={TAPEX: Table pre-training via learning a neural SQL executor},
  author={Liu, Qian and Chen, Bei and Guo, Jiaqi and Ziyadi, Morteza and Lin, Zeqi and Chen, Weizhu and Lou, Jian-Guang},
  journal={arXiv preprint arXiv:2107.07653},
  year={2021}
}

@inproceedings{pasupat2015compositional,
  title={Compositional semantic parsing on semi-structured tables},
  author={Pasupat, Panupong and Liang, Percy},
  booktitle={Proceedings of the 53rd Annual Meeting of the Association for Computational Linguistics and the 7th International Joint Conference on Natural Language Processing (Volume 1: Long Papers)},
  pages={1470--1480},
  year={2015}
}

@article{cheng2022binding,
  title={Binding language models in symbolic languages},
  author={Cheng, Zhoujun and Xie, Tianbao and Shi, Peng and Li, Chengzu and Nadkarni, Rahul and Hu, Yushi and Xiong, Caiming and Radev, Dragomir and Ostendorf, Mari and Zettlemoyer, Luke and others},
  journal={arXiv preprint arXiv:2210.02875},
  year={2022}
}

@article{wang2024chain,
  title={Chain-of-table: Evolving tables in the reasoning chain for table understanding},
  author={Wang, Zilong and Zhang, Hao and Li, Chun-Liang and Eisenschlos, Julian Martin and Perot, Vincent and Wang, Zifeng and Miculicich, Lesly and Fujii, Yasuhisa and Shang, Jingbo and Lee, Chen-Yu and others},
  journal={arXiv preprint arXiv:2401.04398},
  year={2024}
}

@inproceedings{eisenschlos2020understanding,
  title={Understanding tables with intermediate pre-training},
  author={Eisenschlos, Julian and Krichene, Syrine and M{\"u}ller, Thomas},
  booktitle={Findings of the Association for Computational Linguistics: EMNLP 2020},
  pages={281--296},
  year={2020}
}

@inproceedings{yin2020tabert,
  title={TaBERT: Pretraining for joint understanding of textual and tabular data},
  author={Yin, Pengcheng and Neubig, Graham and Yih, Wen-tau and Riedel, Sebastian},
  booktitle={Proceedings of the 58th annual meeting of the association for computational linguistics},
  pages={8413--8426},
  year={2020}
}

@article{zhong2017seq2sql,
  title={Seq2sql: Generating structured queries from natural language using reinforcement learning},
  author={Zhong, Victor and Xiong, Caiming and Socher, Richard},
  journal={arXiv preprint arXiv:1709.00103},
  year={2017}
}

@inproceedings{zhao2024tapera,
  title={TaPERA: Enhancing faithfulness and interpretability in long-form table QA by content planning and execution-based reasoning},
  author={Zhao, Yilun and Chen, Lyuhao and Cohan, Arman and Zhao, Chen},
  booktitle={Proceedings of the 62nd Annual Meeting of the Association for Computational Linguistics (Volume 1: Long Papers)},
  pages={12824--12840},
  year={2024}
}

@article{chen2019tabfact,
  title={Tabfact: A large-scale dataset for table-based fact verification},
  author={Chen, Wenhu and Wang, Hongmin and Chen, Jianshu and Zhang, Yunkai and Wang, Hong and Li, Shiyang and Zhou, Xiyou and Wang, William Yang},
  journal={arXiv preprint arXiv:1909.02164},
  year={2019}
}

@inproceedings{gu2022pasta,
  title={PASTA: table-operations aware fact verification via sentence-table cloze pre-training},
  author={Gu, Zihui and Fan, Ju and Tang, Nan and Nakov, Preslav and Zhao, Xiaoman and Du, Xiaoyong},
  booktitle={Proceedings of the 2022 conference on empirical methods in natural language processing},
  pages={4971--4983},
  year={2022}
}

@inproceedings{schlichtkrull2021joint,
  title={Joint verification and reranking for open fact checking over tables},
  author={Schlichtkrull, Michael and Karpukhin, Vladimir and Oguz, Barlas and Lewis, Mike and Yih, Wen-tau and Riedel, Sebastian},
  booktitle={Proceedings of the 59th Annual Meeting of the Association for Computational Linguistics and the 11th International Joint Conference on Natural Language Processing (Volume 1: Long Papers)},
  pages={6787--6799},
  year={2021}
}

@inproceedings{yang2020program,
  title={Program enhanced fact verification with verbalization and graph attention network},
  author={Yang, Xiaoyu and Nie, Feng and Feng, Yufei and Liu, Quan and Chen, Zhigang and Zhu, Xiaodan},
  booktitle={Proceedings of the 2020 Conference on Empirical Methods in Natural Language Processing (EMNLP)},
  pages={7810--7825},
  year={2020}
}

@inproceedings{asai2023self,
  title={Self-rag: Learning to retrieve, generate, and critique through self-reflection},
  author={Asai, Akari and Wu, Zeqiu and Wang, Yizhong and Sil, Avirup and Hajishirzi, Hannaneh},
  booktitle={The Twelfth International Conference on Learning Representations},
  year={2023}
}

@inproceedings{gao2023enabling,
  title={Enabling large language models to generate text with citations},
  author={Gao, Tianyu and Yen, Howard and Yu, Jiatong and Chen, Danqi},
  booktitle={Proceedings of the 2023 Conference on Empirical Methods in Natural Language Processing},
  pages={6465--6488},
  year={2023}
}

@inproceedings{gao2023rarr,
  title={Rarr: Researching and revising what language models say, using language models},
  author={Gao, Luyu and Dai, Zhuyun and Pasupat, Panupong and Chen, Anthony and Chaganty, Arun Tejasvi and Fan, Yicheng and Zhao, Vincent and Lao, Ni and Lee, Hongrae and Juan, Da-Cheng and others},
  booktitle={Proceedings of the 61st Annual Meeting of the Association for Computational Linguistics (Volume 1: Long Papers)},
  pages={16477--16508},
  year={2023}
}

@article{rashkin2023measuring,
  title={Measuring attribution in natural language generation models},
  author={Rashkin, Hannah and Nikolaev, Vitaly and Lamm, Matthew and Aroyo, Lora and Collins, Michael and Das, Dipanjan and Petrov, Slav and Tomar, Gaurav Singh and Turc, Iulia and Reitter, David},
  journal={Computational Linguistics},
  volume={49},
  pages={777--840},
  year={2023}
}

@article{bohnet2022attributed,
  title={Attributed question answering: Evaluation and modeling for attributed large language models},
  author={Bohnet, Bernd and Tran, Vinh Q and Verga, Pat and Aharoni, Roee and Andor, Daniel and Soares, Livio Baldini and Ciaramita, Massimiliano and Eisenstein, Jacob and Ganchev, Kuzman and Herzig, Jonathan and others},
  journal={arXiv preprint arXiv:2212.08037},
  year={2022}
}

@article{wallat2024correctness,
  title={Correctness is not faithfulness in rag attributions},
  author={Wallat, Jonas and Heuss, Maria and de Rijke, Maarten and Anand, Avishek},
  journal={arXiv preprint arXiv:2412.18004},
  year={2024}
}

@article{wei2022chain,
  title={Chain-of-thought prompting elicits reasoning in large language models},
  author={Wei, Jason and Wang, Xuezhi and Schuurmans, Dale and Bosma, Maarten and Xia, Fei and Chi, Ed and Le, Quoc V and Zhou, Denny and others},
  journal={Advances in neural information processing systems},
  volume={35},
  pages={24824--24837},
  year={2022}
}

@article{ouyang2022training,
  title={Training language models to follow instructions with human feedback},
  author={Ouyang, Long and Wu, Jeffrey and Jiang, Xu and Almeida, Diogo and Wainwright, Carroll and Mishkin, Pamela and Zhang, Chong and Agarwal, Sandhini and Slama, Katarina and Ray, Alex and others},
  journal={Advances in neural information processing systems},
  volume={35},
  pages={27730--27744},
  year={2022}
}

@article{shao2024deepseekmath,
  title={Deepseekmath: Pushing the limits of mathematical reasoning in open language models},
  author={Shao, Zhihong and Wang, Peiyi and Zhu, Qihao and Xu, Runxin and Song, Junxiao and Bi, Xiao and Zhang, Haowei and Zhang, Mingchuan and Li, YK and Wu, Yang and others},
  journal={arXiv preprint arXiv:2402.03300},
  year={2024}
}

@article{guo2025deepseek,
  title={Deepseek-r1: Incentivizing reasoning capability in llms via reinforcement learning},
  author={Guo, Daya and Yang, Dejian and Zhang, Haowei and Song, Junxiao and Wang, Peiyi and Zhu, Qihao and Xu, Runxin and Zhang, Ruoyu and Ma, Shirong and Bi, Xiao and others},
  journal={arXiv preprint arXiv:2501.12948},
  year={2025}
}

@article{rafailov2023direct,
  title={Direct preference optimization: Your language model is secretly a reward model},
  author={Rafailov, Rafael and Sharma, Archit and Mitchell, Eric and Manning, Christopher D and Ermon, Stefano and Finn, Chelsea},
  journal={Advances in neural information processing systems},
  volume={36},
  pages={53728--53741},
  year={2023}
}

@article{schulman2017proximal,
  title={Proximal policy optimization algorithms},
  author={Schulman, John and Wolski, Filip and Dhariwal, Prafulla and Radford, Alec and Klimov, Oleg},
  journal={arXiv preprint arXiv:1707.06347},
  year={2017}
}

@article{dang2025reinforcement,
  title={Reinforcement Learning for Reasoning in Small LLMs: What Works and What Doesn't},
  author={Dang, Quy-Anh and Ngo, Chris},
  journal={arXiv preprint arXiv:2503.16219},
  year={2025}
}

@article{hu2022lora,
  title={Lora: Low-rank adaptation of large language models.},
  author={Hu, Edward J and Shen, Yelong and Wallis, Phillip and Allen-Zhu, Zeyuan and Li, Yuanzhi and Wang, Shean and Wang, Liang and Chen, Weizhu and others},
  journal={Iclr},
  volume={1},
  pages={3},
  year={2022}
}

@article{qwen2024qwen25,
  title={Qwen2.5 Technical Report},
  author={Yang, An and Yang, Baosong and Zhang, Beichen and Hui, Binyuan and Zheng, Bo and Yu, Bowen and Li, Chengyuan and Liu, Dayiheng and Huang, Fei and Wei, Haoran and Lin, Huan and Yang, Jian and Tu, Jianhong and Zhang, Jianwei and Yang, Jianxin and Yang, Jiaxi and Zhou, Jingren and Lin, Junyang and Dang, Kai and Lu, Keming and Bao, Keqin and Yang, Kexin and Yu, Le and Li, Mei and Xue, Mingfeng and Zhang, Pei and Zhu, Qin and Men, Rui and Lin, Runji and Li, Tianhao and Tang, Tianyi and Xia, Tingyu and Ren, Xingzhang and Ren, Xuancheng and Fan, Yang and Su, Yang and Zhang, Yichang and Wan, Yu and Liu, Yuqiong and Cui, Zeyu and Zhang, Zhenru and Qiu, Zihan},
  journal={arXiv preprint arXiv:2412.15115},
  year={2025}
}

@article{llama2024llama3,
  title={The llama 3 herd of models},
  author={Grattafiori, Aaron and Dubey, Abhimanyu and Jauhri, Abhinav and Pandey, Abhinav and Kadian, Abhishek and Al-Dahle, Ahmad and Letman, Aiesha and Mathur, Akhil and Schelten, Alan and Vaughan, Alex and others},
  journal={arXiv preprint arXiv:2407.21783},
  year={2024}
}

@inproceedings{devlin2019bert,
  title={Bert: Pre-training of deep bidirectional transformers for language understanding},
  author={Devlin, Jacob and Chang, Ming-Wei and Lee, Kenton and Toutanova, Kristina},
  booktitle={Proceedings of the 2019 conference of the North American chapter of the association for computational linguistics: human language technologies, volume 1 (long and short papers)},
  pages={4171--4186},
  year={2019}
}

@article{nan2022fetaqa,
  title={FeTaQA: Free-form table question answering},
  author={Nan, Linyong and Hsieh, Chiachun and Mao, Ziming and Lin, Xi Victoria and Verma, Neha and Zhang, Rui and Kry{\'s}ci{\'n}ski, Wojciech and Schoelkopf, Hailey and Kong, Riley and Tang, Xiangru and others},
  journal={Transactions of the Association for Computational Linguistics},
  volume={10},
  pages={35--49},
  year={2022},
  publisher={MIT Press One Broadway, 12th Floor, Cambridge, Massachusetts 02142, USA~…}
}

@article{he2020deberta,
  title={Deberta: Decoding-enhanced bert with disentangled attention},
  author={He, Pengcheng and Liu, Xiaodong and Gao, Jianfeng and Chen, Weizhu},
  journal={arXiv preprint arXiv:2006.03654},
  year={2020}
}

\clearpage
\appendix

\section{Data Construction Details}
\label{app:data}
 
\paragraph{Gold trace generation.}
Each training example is generated by prompting Claude Opus 4.5 with a system prompt specifying the JSON format, followed by the linearized table and question. The system prompt instructs the model to produce 3--4 reasoning steps, each citing 1--4 specific cells using $[\text{row}_i, \text{col}_j]$ coordinates.
 
\paragraph{Verification pipeline.}
Generated traces undergo three automated checks: (1) JSON parse validity; (2) all cited cell coordinates are within table bounds; (3) step count is between 3--4. Failed examples are batched and re-prompted with error-specific instructions. Final repair rate: WTQ required 18 repair batches; FeTaQA and TabFact were 100\% valid on first generation.

\section{Training Hyperparameters}
\label{app:hyperparams}
 
\begin{table}[ht]
\centering
\caption{Hyperparameters for SFT and GRPO training across all model scales.}
\label{tab:hyperparams}
\small
\setlength{\tabcolsep}{3pt}
\begin{tabular}{@{}l cc cc@{}}
\toprule
& \multicolumn{2}{c}{\textbf{SFT}} & \multicolumn{2}{c}{\textbf{GRPO}} \\
\cmidrule(lr){2-3} \cmidrule(lr){4-5}
\textbf{Parameter} & \textbf{1--3B} & \textbf{7--8B} & \textbf{1--3B} & \textbf{7--8B} \\
\midrule
LoRA rank $r$ & 32 & 64 & 16 & 32 \\
LoRA $\alpha$ & 64 & 128 & 32 & 64 \\
LoRA dropout & 0.05 & 0.05 & 0.05 & 0.05 \\
Learning rate & 2e-4 & 2e-4 & 5e-5 & 5e-5 \\
Batch size & 4 & 8 & 2 & 4 \\
Grad.\ accum. & 4 & 2 & 8 & 4 \\
Eff.\ batch size & 16 & 16 & 16 & 16 \\
Epochs & 3 & 3 & 1 & 1 \\
Scheduler & \multicolumn{2}{c}{cosine} & \multicolumn{2}{c}{cosine} \\
Precision & \multicolumn{2}{c}{bf16} & \multicolumn{2}{c}{bf16} \\
Max seq.\ length & \multicolumn{2}{c}{2048} & \multicolumn{2}{c}{1024} \\
Group size $G$ & \multicolumn{2}{c}{---} & \multicolumn{2}{c}{8} \\
Temperature & \multicolumn{2}{c}{---} & \multicolumn{2}{c}{0.9} \\
\bottomrule
\end{tabular}
\end{table}

\section{Reward Component Definitions}
\label{app:reward}

Full mathematical definitions for the composite reward (Section~\ref{sec:reward}).
 
\paragraph{Answer F1.}
Given predicted answer tokens $P$ and gold answer tokens $G$:
\begin{equation}
\text{Prec} = \frac{|P \cap G|}{|P|}, \quad \text{Rec} = \frac{|P \cap G|}{|G|}
\end{equation}
\begin{equation}
R_{\text{ans}} = \frac{2 \cdot \text{Prec} \cdot \text{Rec}}{\text{Prec} + \text{Rec}}
\end{equation}
When multiple gold answers exist, we take the maximum F1.
 
\paragraph{Citation validity.}
For output with $N$ total cited cells across all steps:
\begin{equation}
R_{\text{cite}} = \frac{1}{N} \sum_{i=1}^{N} \mathbf{1}[0 \leq r_i < R \wedge 0 \leq c_i < C]
\end{equation}
 
\paragraph{Faithfulness.}
For each reasoning step $s_k$ with cited cells $\{(r_i, c_j)\}$, we concatenate the cell values into evidence string $e_k$, then compute:
\begin{equation}
R_{\text{faith}} = \frac{1}{K} \sum_{k=1}^{K} \text{NLI}_{\text{entail}}(e_k, s_k)
\end{equation}
where $\text{NLI}_{\text{entail}}$ is the entailment probability from DeBERTa-v3-base (\texttt{cross-encoder/nli-deberta-v3-base}).
Steps with no cited cells receive score 0.
 
\paragraph{Parsimony.}
For each step with $n_k$ cited cells:
\begin{equation}
R_{\text{pars}} = \frac{1}{K} \sum_{k=1}^{K}
\begin{cases}
1.0 & n_k \leq 3 \\
\frac{8 - n_k}{5} & 3 < n_k < 8 \\
0.0 & n_k \geq 8
\end{cases}
\end{equation}

\section{Ablation Visualization}
\label{app:ablation_viz}
 
\begin{figure}[ht]
\centering
\includegraphics[width=\columnwidth]{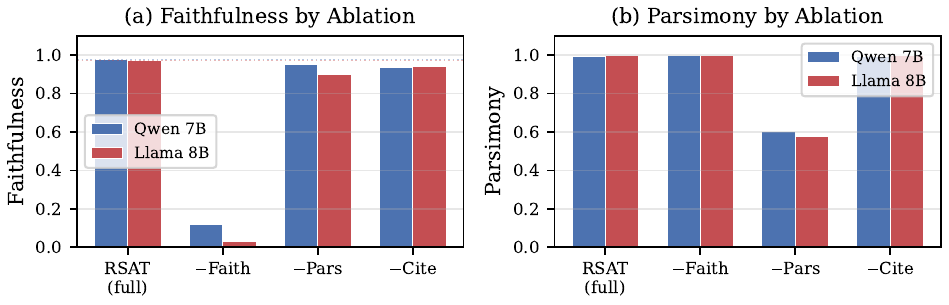}
\caption{\textbf{Ablation impact on faithfulness and parsimony.} Removing faithfulness reward causes near-total collapse; removing parsimony causes significant over-citation.}
\label{fig:ablation_bars}
\end{figure}

Figure~\ref{fig:ablation_bars} visualizes the ablation study results from Section~\ref{sec:results}, showing the effect of removing each reward component on faithfulness and parsimony.
 
\section{Output Format Specification}
\label{app:format}
 
The RSAT output format is a JSON object with the following schema:
 
\begin{lstlisting}[language={}]
{
  "reasoning_steps": [
    {
      "step": "<natural language claim>",
      "cited_cells": [
        [<row_index>, <col_index>],
        ...
      ]
    },
    ...
  ],
  "answer": "<final answer string>"
}
\end{lstlisting}

\section{Training Curves}
\label{app:training_curves}
 
\begin{figure}[ht]
\centering
\includegraphics[width=\columnwidth]{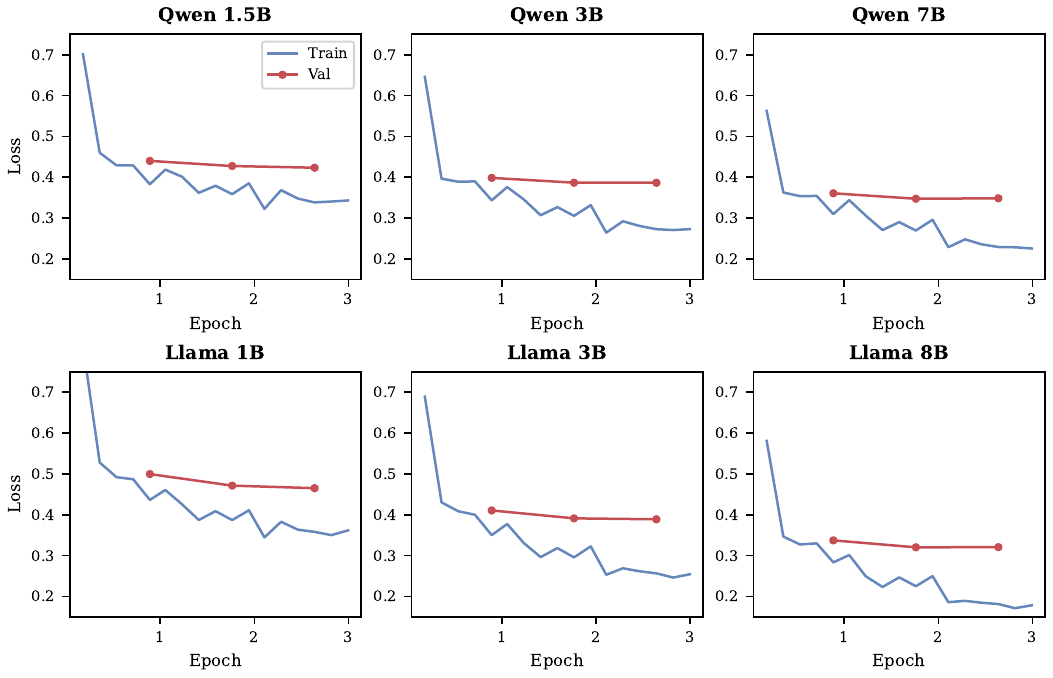}
\caption{\textbf{SFT training curves.} Train and validation loss over 3 epochs for all 6 models. All models converge without overfitting.}
\label{fig:sft_curves}
\end{figure}

\paragraph{SFT convergence.}
Figure~\ref{fig:sft_curves} plots training and validation loss for all six models during SFT. All models converge smoothly with no significant overfitting: validation loss plateaus after epoch~1 while training loss continues to decrease slightly. The Llama models start at higher loss (${\sim}$0.70) compared to Qwen (${\sim}$0.56--0.70) but converge to comparable final values.
 
\begin{figure}[ht]
\centering
\includegraphics[width=\columnwidth]{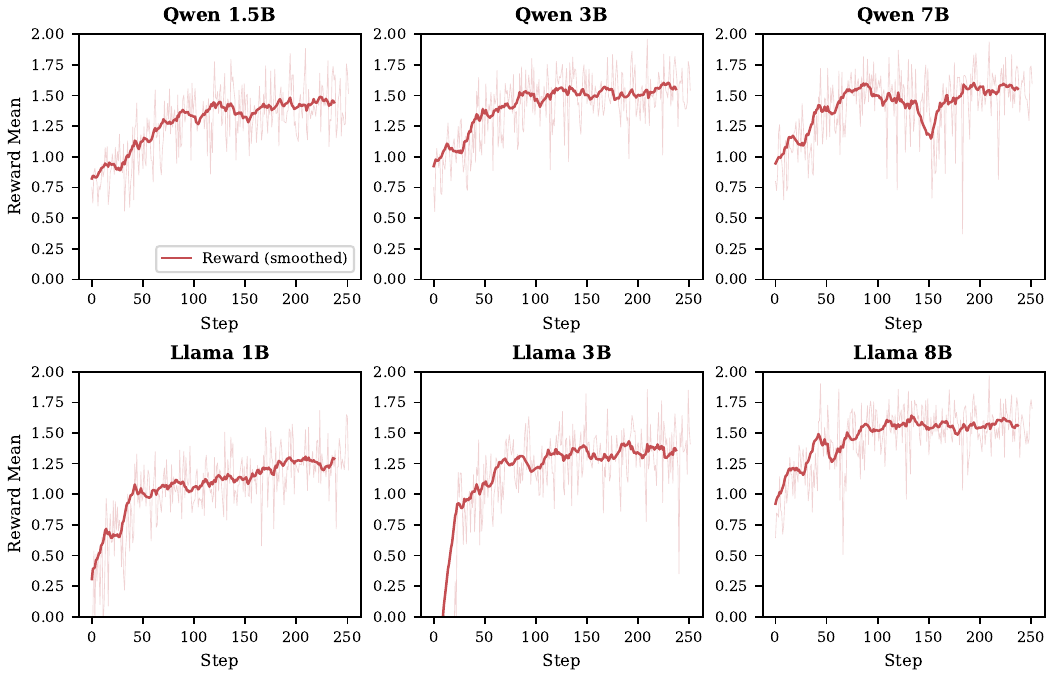}
\caption{\textbf{GRPO reward curves.} Composite reward mean increases for all models despite near-zero loss. Smoothed with a 15-step moving average; raw values shown in light shading.}
\label{fig:grpo_curves}
\end{figure}

\paragraph{GRPO reward dynamics.}
Figure~\ref{fig:grpo_curves} shows the composite reward mean during GRPO training (250 steps). All models show a clear upward trend from ${\sim}$0.75 to ${\sim}$1.5+, confirming that GRPO successfully optimizes the reward signal. The near-zero GRPO loss is expected and not indicative of a training failure.
 
\section{GRPO Training Statistics}
\label{app:grpo_stats}
 
Table~\ref{tab:grpo_stats} reports per-model GRPO training statistics.
 
\begin{table}[ht]
\centering
\caption{GRPO training statistics per model (250 steps each).}
\label{tab:grpo_stats}
\small
\setlength{\tabcolsep}{4pt}
\begin{tabular}{@{}l ccc@{}}
\toprule
\textbf{Model} & \textbf{Time (min)} & \textbf{Final Reward} & \textbf{Final Entropy} \\
\midrule
Qwen 1.5B & 120 & 1.560 & 0.118 \\
Qwen 3B   & 116 & 1.623 & 0.057 \\
Qwen 7B   & 265 & 1.590 & 0.038 \\
Llama 1B  & 105 & 1.459 & 0.197 \\
Llama 3B  & 143 & 1.488 & 0.056 \\
Llama 8B  & 123 & 1.660 & 0.033 \\
\bottomrule
\end{tabular}
\end{table}
 
\noindent The final entropy values (0.03--0.20) confirm that GRPO sharpens the policy distribution: smaller entropy indicates more confident, focused generations. Llama~1B retains the highest entropy (0.197), consistent with its lower faithfulness (0.480). Llama~8B achieves the highest final reward (1.660), aligning with its top faithfulness score (0.972).

\section{Compute Budget}
\label{app:compute}
 
\begin{table}[ht]
\centering
\caption{Wall-clock time on a single NVIDIA H100 80GB GPU.}
\label{tab:compute}
\small
\setlength{\tabcolsep}{3.5pt}
\begin{tabular}{@{}l rrrr@{}}
\toprule
\textbf{Model} & \textbf{SFT} & \textbf{GRPO} & \textbf{Eval} & \textbf{Total} \\
\midrule
Qwen 1.5B &   7 min & 120 min &  26 min & 153 min \\
Qwen 3B   &  11 min & 116 min &  20 min & 147 min \\
Qwen 7B   &  16 min & 265 min &  51 min & 333 min \\
Llama 1B  &   5 min & 105 min &  27 min & 137 min \\
Llama 3B  &  11 min & 143 min &  25 min & 179 min \\
Llama 8B  &  18 min & 123 min &  16 min & 158 min \\
\midrule
\textbf{6-model subtotal} & \textbf{68 min} & \textbf{874 min} & \textbf{165 min} & \textbf{18.4 hrs} \\
\midrule
\multicolumn{4}{@{}l}{\textit{Ablation runs (6 configs $\times$ train + eval)}} & 18.3 hrs \\
\midrule
\textbf{Grand Total} & \multicolumn{3}{c}{} & \textbf{36.8 hrs} \\
\bottomrule
\end{tabular}
\end{table}
 
\noindent GRPO dominates training time due to generating $G{=}8$ candidates per question and computing per-step NLI faithfulness scores. Qwen~7B is the most expensive model (265~min GRPO) because its longer completions require more generation and scoring time.

\section{Completion Length Analysis}
\label{app:completion_length}
 
\begin{figure}[ht]
\centering
\includegraphics[width=\columnwidth]{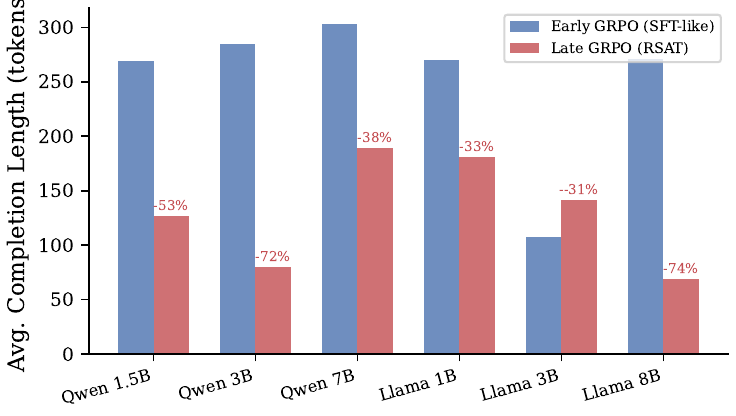}
\caption{\textbf{Completion length reduction during GRPO.} Early GRPO completions (SFT-like behavior) vs.\ late GRPO completions (RSAT behavior). All models learn to produce shorter, more focused outputs.}
\label{fig:completion_length}
\end{figure}

Figure~\ref{fig:completion_length} compares average completion lengths at the start and end of GRPO training. All models show substantial compression: the average reduction is 40--55\% across models. This confirms that the parsimony reward teaches models to produce concise, focused reasoning rather than verbose multi-step chains. The effect is especially pronounced for the 3B and 7--8B models, which begin with longer completions due to greater generation capacity.
 
\section{Post-hoc Failure Analysis}
\label{app:posthoc_analysis}
 
Table~\ref{tab:posthoc_failure} categorizes the failure modes of post-hoc attribution. The two-pass approach asks the model to first generate a chain-of-thought answer, then retroactively map its reasoning to cell coordinates in JSON format. The dominant failure mode is \emph{empty or non-JSON output} (76--100\% of examples), not JSON parse errors. The models simply do not produce the required format in the second pass---they either repeat the question, generate a natural language response, or produce truncated output. This suggests the failure is fundamental: small models lack the working memory to re-read a table and retro-fit structured citations.
 
\begin{table}[ht]
\centering
\caption{Post-hoc failure modes (500 test examples per model).}
\label{tab:posthoc_failure}
\small
\setlength{\tabcolsep}{4pt}
\begin{tabular}{@{}l ccc@{}}
\toprule
\textbf{Model} & \textbf{Fmt\%} & \textbf{Empty/No-JSON} & \textbf{Valid} \\
\midrule
Qwen 1.5B & 24.4\% & 75.6\% & 24.4\% \\
Qwen 3B   &  0.4\% & 99.6\% &  0.4\% \\
Qwen 7B   & 14.0\% & 86.0\% & 14.0\% \\
Llama 1B  & 19.8\% & 80.2\% & 19.8\% \\
Llama 3B  &  9.0\% & 91.0\% &  9.0\% \\
Llama 8B  &  4.0\% & 96.0\% &  4.0\% \\
\midrule
\textbf{Average} & \textbf{12.0\%} & \textbf{88.1\%} & \textbf{12.0\%} \\
\bottomrule
\end{tabular}
\end{table}
 
\noindent Counterintuitively, larger models perform \emph{worse}: Llama~8B (4.0\%) is below Llama~1B (19.8\%). Larger models produce longer free-text responses in the second pass, making JSON compliance harder.

\section{Faithfulness Improvement Visualization}
\label{app:faithfulness_viz}
 
\begin{figure}[ht]
\centering
\includegraphics[width=\columnwidth]{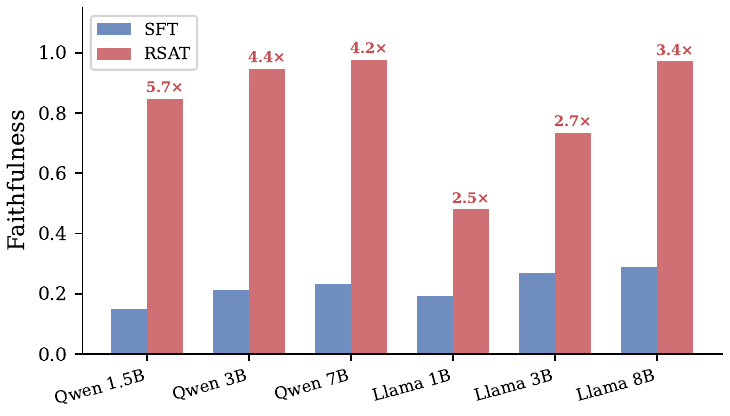}
\caption{\textbf{Faithfulness: SFT vs.\ RSAT per model.} All models show substantial gains; the 3.7$\times$ average is driven by near-ceiling Qwen performance.}
\label{fig:faith_waterfall}
\end{figure}

Figure~\ref{fig:faith_waterfall} visualizes the faithfulness improvement from SFT to RSAT for each model. Qwen~1.5B achieves the largest relative gain (5.7$\times$), while Llama~1B shows the smallest absolute improvement---still a 2.5$\times$ gain. Both architecture families benefit substantially from GRPO, confirming that the faithfulness reward transfers across model families.
 
\end{document}